\documentclass[conference]{IEEEtran}
\IEEEoverridecommandlockouts
\usepackage{cite}
\usepackage{mdframed}
\usepackage{amsmath,amssymb,amsfonts}
\usepackage{algorithmic}
\usepackage{graphicx}

\usepackage{textcomp}
\usepackage[table,xcdraw]{xcolor}

\usepackage{multirow}
\usepackage[table,xcdraw]{xcolor}
\usepackage[utf8]{inputenc}
\usepackage{float}
\usepackage[colorinlistoftodos]{todonotes}
\usepackage{caption}
\usepackage{subcaption}
\usepackage{comment}
\usepackage{threeparttable}
\input{./comments.tex}
\def\BibTeX{{\rm B\kern-.05em{\sc i\kern-.025em b}\kern-.08em
    T\kern-.1667em\lower.7ex\hbox{E}\kern-.125emX}}
    
\begin{document}

\title{AUTHENTICATION \\ \LARGE Identifying Rare Failure Modes in Autonomous Vehicle Perception Systems using Adversarially Guided Diffusion Models

\thanks{Approved for Public Release; Distribution Unlimited. Public Release Case Number 25-0077.  \\©2025 The MITRE Corporation. ALL RIGHTS RESERVED.

}
}

\author{
\IEEEauthorblockN{1\textsuperscript{st} Mohammad Zarei}
\IEEEauthorblockA{\textit{MITRE CORP.}\\
San Diego, USA \\
mzarei@mitre.org}
\and
\IEEEauthorblockN{2\textsuperscript{nd} Melanie A Jutras}
\IEEEauthorblockA{\textit{MITRE CORP.}\\
Bedford, USA \\
mjutras@mitre.org}
\and
\IEEEauthorblockN{3\textsuperscript{rd} Eliana Evans}
\IEEEauthorblockA{\textit{MITRE CORP.}\\
Austin, USA\\
annaevans@mitre.org}
\and
\IEEEauthorblockN{4\textsuperscript{th} Mike Tan}
\IEEEauthorblockA{
\textit{Cranium} \\
San Diego , USA \\
mtan@cranium.ai}
\and
\IEEEauthorblockN{5\textsuperscript{th} Omid Aaramoon}
\IEEEauthorblockA{
\textit{Booz Allen Hamilton}\\
Virginia, USA \\
aramoon\_omid@bah.com}
}

\maketitle

\begin{abstract}
Autonomous Vehicles (AVs) rely on artificial intelligence (AI) to accurately detect objects and interpret their surroundings. However, even when trained using millions of miles of real-world data, AVs are often unable to detect rare failure modes (RFMs). The problem of RFMs is commonly referred to as the “long-tail challenge”, due to the distribution of data including many instances that are very rarely seen. In this paper, we present a novel approach that utilizes advanced generative and explainable AI techniques to aid in understanding RFMs. Our methods can be used to enhance the robustness and reliability of AVs when combined with both downstream model training and testing. We extract segmentation masks for objects of interest (e.g., cars) and invert them to create environmental masks. These masks, combined with carefully crafted text prompts, are fed into a custom diffusion model. We leverage the Stable Diffusion inpainting model guided by adversarial noise optimization to generate images containing diverse environments designed to evade object detection models and expose vulnerabilities in AI systems. Finally, we produce natural language descriptions of the generated RFMs that can guide developers and policymakers to improve the safety and reliability of AV systems.
\end{abstract}


\section{Introduction}

There has been significant progress in the development of autonomous vehicles (AVs) in recent years due to rapid advancements in artificial intelligence (AI). However, critical functionality challenges must be addressed before AVs are considered reliable and robust. Due to the safety critical nature of AVs, as well as the potential for great impact on the broader transportation industry and society, both government and non-government organizations have released best practices and standards. In 2022, the International Organization for Standardization (ISO) introduced Safety of the Intended Functionality (SOTIF) as a safety requirements standard in response to the unique challenges and limitations of AV AI systems \cite{WANG202417}. One challenge SOTIF addresses is functional deficiencies in AV perception systems including object detection, classification, and tracking models traditionally trained on real-world data. Because AV perception systems traditionally rely on real-world data collection for training, they fail to include the full range of environmental conditions such as diverse weather patterns, varied lighting, or complex urban environments. As a result, AI models trained and tested on these datasets may perform well in ideal conditions but fail to generalize under rare or unexpected situations, resulting in object detection errors such as missed objects, misclassifications, or hallucinations. These RFMs occur infrequently but can have severe consequences, a problem referred to as the long-tail challenge \cite{Wang_LongTail2022}. Comprehensive testing methods that systematically explore the long-tail of RFMs are crucial for safe AV deployment.

In this paper, we propose a novel pipeline that includes adversarial machine learning, generative diffusion, and explainability techniques to identify and understand RFMs in AV perception systems. We use the Stable Diffusion inpainting model\cite{sd2.0_paper}, a state-of-the-art generative model capable of generating realistic images along with adversarial noise optimization that iteratively refines the initial noise used in the diffusion model. This process is guided by feedback from an object detection model, such as Faster R-CNN \cite{fasterrcnn}, to generate adversarial examples that evade detection. By optimizing the noise based on object detection loss gradients, our method uncovers specific weaknesses in AI perception models that are not apparent under standard testing conditions. These insights can be used improve the robustness of downstream model development. We develop a model-agnostic tool to make our findings accessible to developers, regulators, and stakeholders by generating natural language descriptions that describe the potential RFM causes.

Our method addresses the lack of generalizability in current AV systems as well as explainability of rare failures, resulting in improved safety and transparency. This approach aligns with the U.S. Department of Transportation's vision for safe and transparent AV systems \cite{usdotav_website}. The insights gained from our framework can inform developers and policymakers, and contribute to the development of robust AV systems and effective regulations for AV safety.
 
\section{Related Work}

Long-tail distributions have been studied in the field of statistics broadly with respect to probability distributions, where the central part of the distribution contains the most commonly occurring data and the tail contains a large number of rarely occurring samples. This type of distribution is common in a variety of domains, but especially seen in real world vision data \cite{Wang_LongTail2022}. With respect to autonomous vehicles and rare failures, the long-tail distribution contains a small number of events that are very common (such as our seed test case image), and a large number of events that are rare (the space of various environmental conditions is quite large, although any one of those occurring might be rare). There is increased interest in researching rare events in autonomous vehicle detection systems, due to the potential for serious consequences of misclassified or undetected events we refer to as RFMs. Wang et al. introduce Long-Tail Regularization (LoTR) to mitigate the effects of rare scenarios encountered in autonomous vehicles \cite{Wang_LongTail2022}. 
Liu and Feng \cite{Liu2024} introduce the concept of the curse of rarity (CoR), which de-emphasizes the importance of the distribution having a long-tail, and emphasizes the rare and complex conditions leading to safety-critical failures. These conditions present a challenge for deep learning models to learn solely based on training data. 
In addition, Diffusion Probabilistic Models \cite{ho2020denoisingdiffusionprobabilisticmodels} have become critical and impactful for generating images from noise, offering superior quality, stability, and training simplicity compared to generative adversarial networks (GANs). These qualities form the basis of our guided diffusion model.

To combat this challenge, researchers have focused on generating realistic adversarial images, such as AdvDiffusers \cite{advdiffuser_paper}, which demonstrate the ability of diffusion models to generate realistic images that tend to fool object detection (OD) models. Our Authentication framework extends this approach by generating environmental backgrounds that ensure realistic, targeted attacks within the data manifold.

\section{Methodology}\label{sec:methodology}

We leverage generative diffusion inpainting models and adversarial machine learning techniques to identify and understand RFMs of AV perception systems. By integrating these models with adversarial noise optimization, using consistency verification, multi-modal outputs, and explainability methods, we create a comprehensive framework for exploring the vulnerabilities of AV perception systems. 

The process starts with segmenting the object of interest (e.g., a car). We reverse the mask to extract the environment mask (see Section \ref{SeedImageSegmentation}). Using our adversarially guided diffusion model, along with the environment mask, a prompt, and the original test case, we generate different environments around the object of interest to create RFMs (see Section \ref{RFMs}).

Next, we validate the generated RFMs through our consistency verification process (see Section \ref{Consistency_Verification}) to ensure consistency across modifications, such as altering the color of the object of interest or converting to video to assess the persistence of the RFM (see Section \ref{multi_modal}).

\begin{figure*}
\centering
\includegraphics[width=0.9\textwidth]{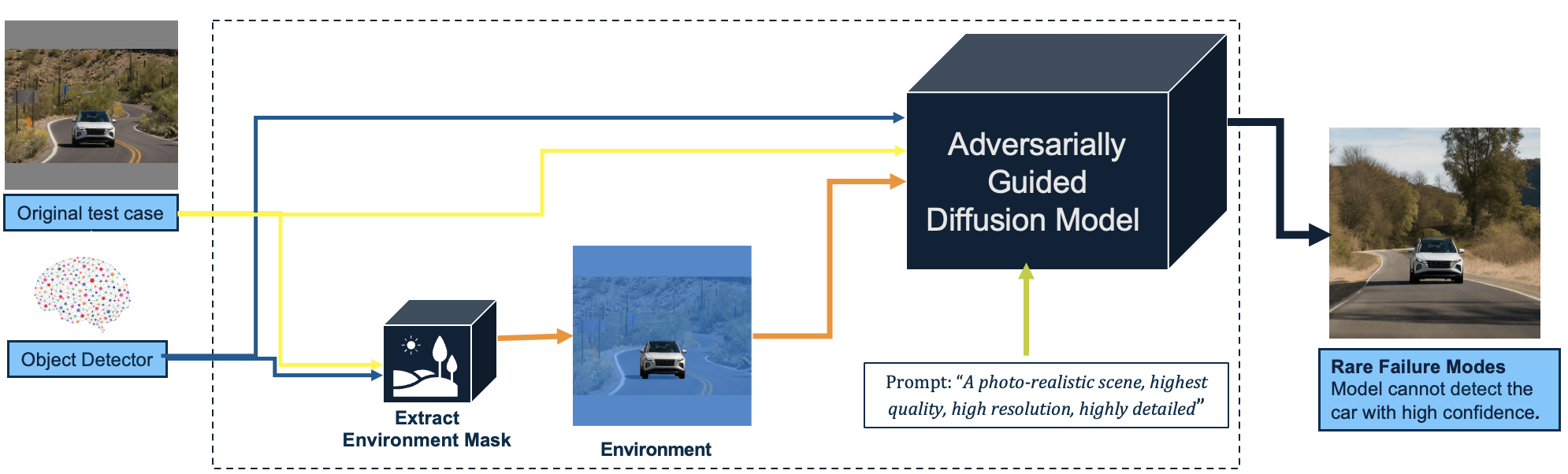}
\caption{Authentication Pipeline: We use the Segment Anything Model \cite{kirillov2023segment} to extract the environment mask around the car. The mask of the environment, along with a prompt, the original image, and the object detector model, is then fed into our adversarially guided diffusion model to generate RFMs.}
\label{auth_pipeline}
\end{figure*}

Finally, we use saliency maps and natural language processing techniques to caption the root cause of the RFMs (see Section \ref{Caption_Generation}). See \figurename \ref{auth_pipeline}, which illustrates the complete pipeline.

\subsection{Seed Image Segmentation}
\label{SeedImageSegmentation}
We first segment the background of a seed image from objects of interest using the Segment Anything Model (SAM) \cite{segment_anything_paper}. Once the segmentation mask of the object of interest is obtained, it is inverted to create the environment mask, which represents all parts of the image that do not contain the objects of interest. The environment mask will be replaced by the inpainting model while objects of interest remain unchanged, ensuring that any detection errors are due to environmental change rather than a modification of the object itself.

\subsection{Rare Failure Mode (RFM) Generation}
\label{RFMs}
We use an adversarially guided inpainting diffusion model to generate new environments while leaving the objects of interest untouched. We use both Stable Diffusion 2.0 and Stable Diffusion XL (SDXL) because of their capability to produce high-quality and contextually coherent images. SDXL is able to generate more highly-detailed images compared to the lighter-weight Stable Diffusion 2.0\cite{sd2.0_paper}\cite{sdxl_paper}. These diffusion models generate images by refining an initial noise tensor through a series of denoising steps guided by a text prompt to produce a coherent environment around the objects of interest \cite{latent_diffusion_models_paper}. To encourage the diffusion models to produce RFM images, we incorporate an adversarial noise optimization process. At each denoising step, we adjust the output based on the gradient of the object detection loss with respect to the noise. This process generates images that are likely to cause object detection errors while maintaining realism  \cite{ho2020denoisingdiffusionprobabilisticmodels}\cite{sd-nae_paper}\cite{diffusion_models_counterfactual_explanations_paper} \cite{rombach2022highresolutionimagesynthesislatent}.

The optimization can be expressed mathematically as:
\[
\boldsymbol{\epsilon}_{\text{optimized}} = \boldsymbol{\epsilon} + \alpha \cdot \frac{\partial \mathcal{L}_{\text{od}}}{\partial \mathbf{z}_t} \cdot \frac{\partial \mathbf{z}_t}{\partial \boldsymbol{\epsilon}}
\]
where \(\mathcal{L}_{\text{od}}\) represents the object detection loss, including classification and bounding box regression. The term \(\frac{\partial \mathcal{L}_{\text{od}}}{\partial \mathbf{z}_t}\) is the gradient of the object detection loss with respect to the latent variable, while \(\frac{\partial \mathbf{z}_t}{\partial \boldsymbol{\epsilon}}\) denotes the gradient of the latent variables with respect to the noise (\(\boldsymbol{\epsilon}\)). Finally, \(\alpha\) is the step size for gradient ascent.

\subsection{Multi-Modal Extension}
\label{multi_modal}
Our primary focus is generating RFM images for image-based object detection models, however, AV perception systems can rely on a combination of modalities. Our framework can generate multi-modal RFM examples including video and LiDAR data, allowing a comprehensive evaluation of AV perception systems in diverse and realistic conditions.

\subsection{Consistency Verification}
\label{Consistency_Verification}
We ensure that generated RFM examples are realistic and effective through a consistency verification step. This step filters out examples that incorporate highly unrealistic scenarios and supports the remaining RFMs as robust and generalizable. We compute object detection, perceptual similarity, and pixel-level similarity metrics. For object detection metrics, we compute confidence, or the confidence of the car detection in the image, and fooling rate, or the percentage of generated images that cause object detection failures. We use these metrics to describe the object detector's performance under difficult environmental conditions. Structural similarity index measure (SSIM) quantifies image similarity based on content, luminance, and contrast. Learned perceptual image patch similarity (LPIPS) measures perceptual similarity based on the difference between high-level image features captured in deep neural network activations. SSIM and LPIPS measure visual and perceptual image qualities, respectively, and make up our perceptual similarity metrics. We use peak signal to noise ratio (PSNR), or the ratio of signal strength between the image and any distorting noise, and mean squared error (MSE), or the average squared difference between image pixel values, as our pixel-level similarity measures.

\subsection{Caption Generation}
\label{Caption_Generation}
We generate captions for RFM images to explain potential causes for object detection errors, making results more accessible to users. We generate image captions with GPT4-o as our image-to-text model due to its availability and state-of-the-art performance on visual perception benchmarks \cite{gpt-4o_blog_post}. The image-to-text model is given input of the thermal image along with system and user text prompts encouraging it to explain the image content and detection performance. To create thermal images, we overlay gradient-weighted class activation maps (Grad-CAM) over the original image. The resulting Grad-CAM images highlight the regions of the image that were important for object detection \cite{grad-cam_paper}. By providing the image-to-text model with a heatmap of the detection model's attention, the captions can explain the "thought process" of the object detection model.

\section{Experimentation}\label{sec:experimentation}

We generate RFM images, alter image elements, and generate RFM videos and explainable captions. These experiments ensure that our results are realistic, generalizable, and accessible.

\subsection{Creating RFMs}
To capture a variety of realistic environmental conditions that present unique object detection challenges for AV perception systems, we present our adversarially-guided inpainting diffusion model with image masks and a text prompt.

\begin{figure}
\centering
\includegraphics[width=\linewidth]{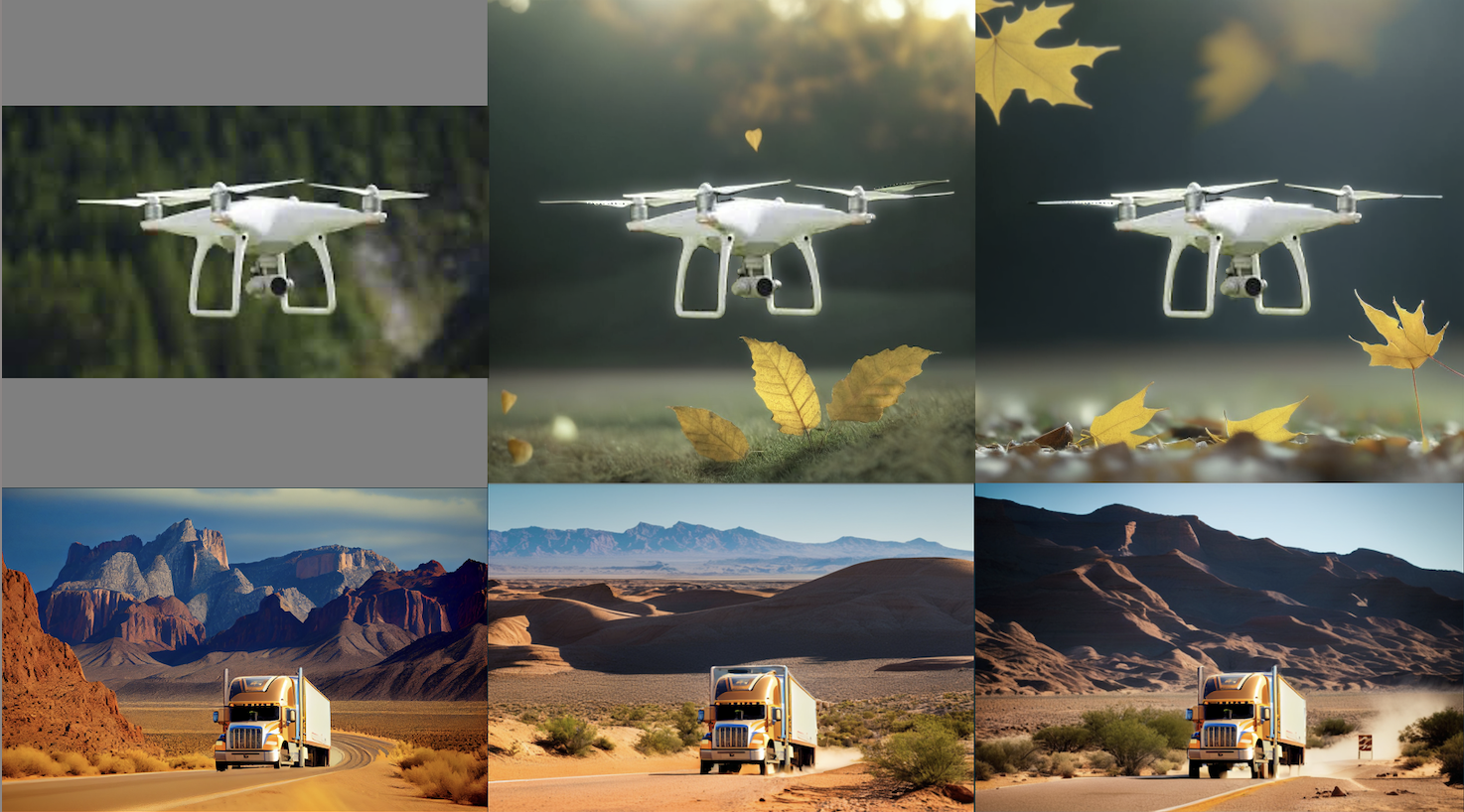}
\caption{Drone and truck seed images on the left, generated RFM images in center and right.}
\label{fig:drone_truck_example}
\end{figure}

\begin{figure}
\centering
\includegraphics[width=\linewidth]{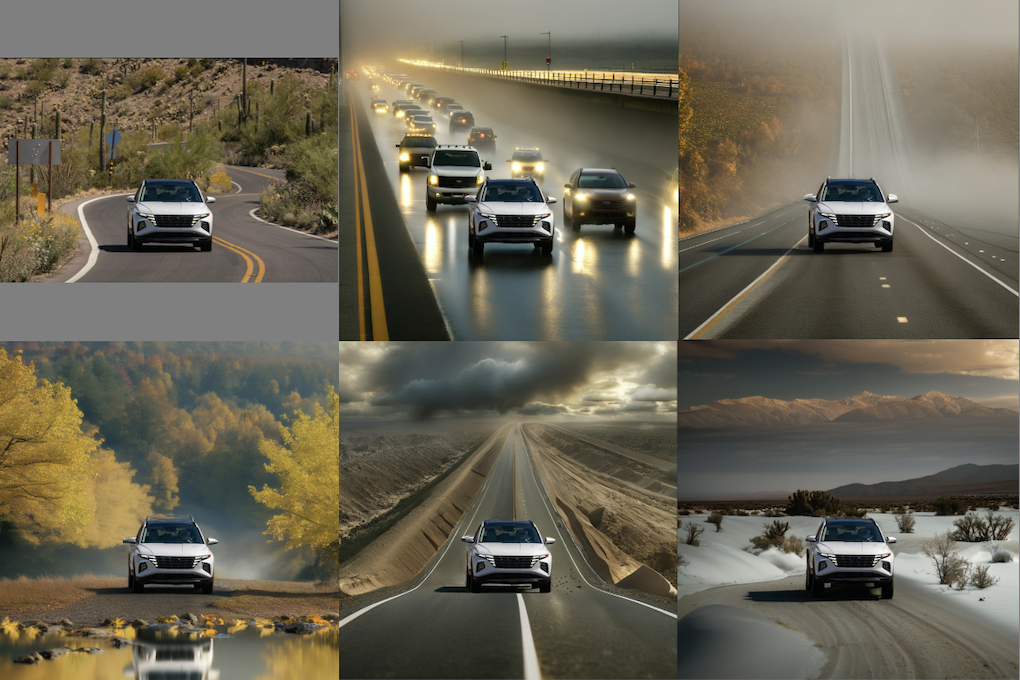}
\caption{Generated images with environmental conditions of night and glare, fog, reflections and yellow foliage, overcast sky, and snow from left to right with seed image in top left.}
\label{fig:sdxl_environments}
\end{figure}

We describe the type of scene we want the model to generate with our positive prompt, for example "high-res, photo-realistic autumn scene with yellow foliage, clear/foggy skies". With the goal of revealing conditions that cause object detection failures, the environment description included in the positive prompt included fog, low light, glaring sun, rain, snow, night, dust storms, reflections, heavy clouds, and wind. Our objects of interest included a white car, a semi truck, and a drone.We discourage the model from generating unrealistic or low-quality images by using a consistent negative prompt of "low res, low resolution, poor quality, bad quality, worst quality, low quality, illustration, 3d, 2d, painting, cartoons, sketch". \figurename \ref{fig:drone_truck_example} shows example drone and truck images. \figurename  \ref{fig:sdxl_environments} shows examples of SDXL generated car images for different environment prompts.

\subsection{Null Text Inversion}
Null text inversion allows us to access the latent space of an image. By doing so, we can change specific features without affecting the rest of the image \cite{null-text_inversion_paper}. We hypothesize that we can change specific image features in a RFM image using null text inversion while maintaining object detection failures, allowing generation of more targeted diverse RFM images.

\begin{figure}
\centering
\includegraphics[width=\linewidth]{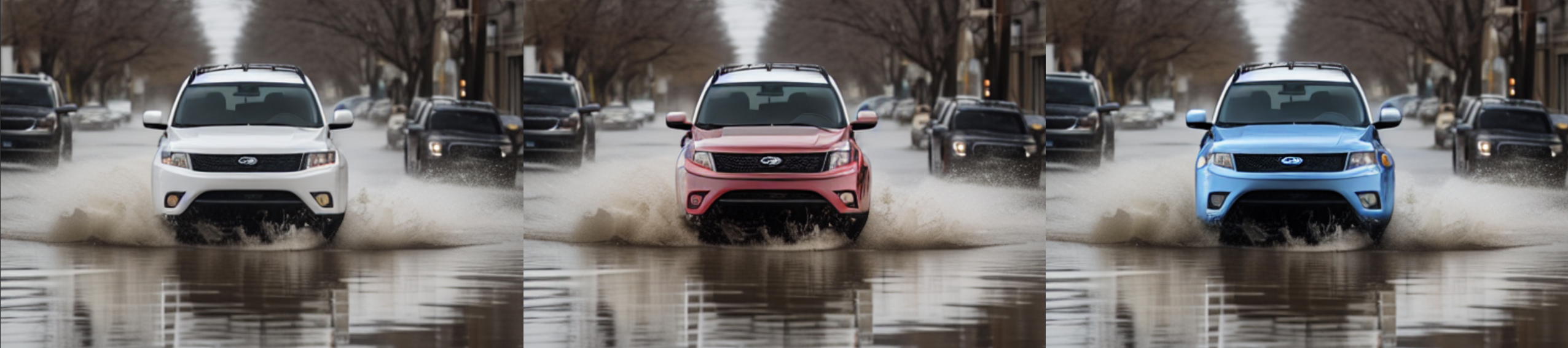}
\caption{RFM status maintained when original image (left) is altered by changing car color to red (center) or blue (right).}
\label{fig:nulltextinversion_car_example}
\end{figure}

\begin{figure}
\centering
\includegraphics[width=\linewidth*2/3]{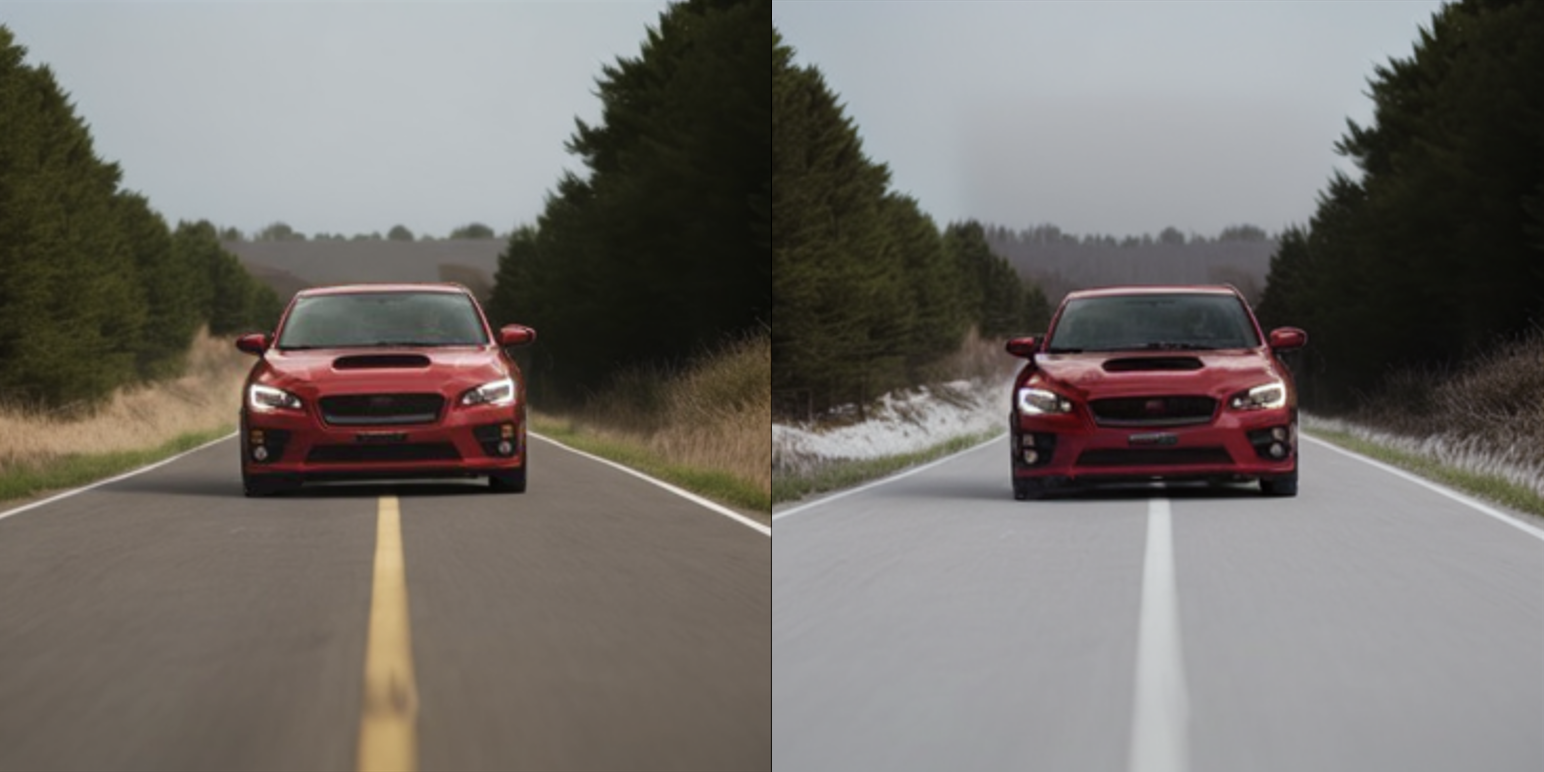}
\caption{RFM status maintained when original image (left) is altered by changing season to winter (right).}
\label{fig:nulltextinversion_road_example}
\end{figure}

We find that RFM status persists after altering the object of interest (\figurename \ref{fig:nulltextinversion_car_example}) or surroundings (\figurename \ref{fig:nulltextinversion_road_example}) using null text inversion. This method could be used to alter RFM images to generate specific environments and test RFM cause hypotheses.

\subsection{Image-to-Video}
By translating RFM images to videos, we can capture how detection failures may occur over time, providing a more realistic simulation for real-world scenarios. We use ConsistI2V, which generates highly consistent videos in appearance, motion, subject, and background, as our image-to-video model. ConsistI2V is a diffusion-based method that uses the first video frame as an anchor through spatiotemporal conditioning and noise initialization to improve spatial, motion, and layout consistency \cite{consistI2V_paper}. 

\begin{figure}
    \centering
    \includegraphics[width=\linewidth]{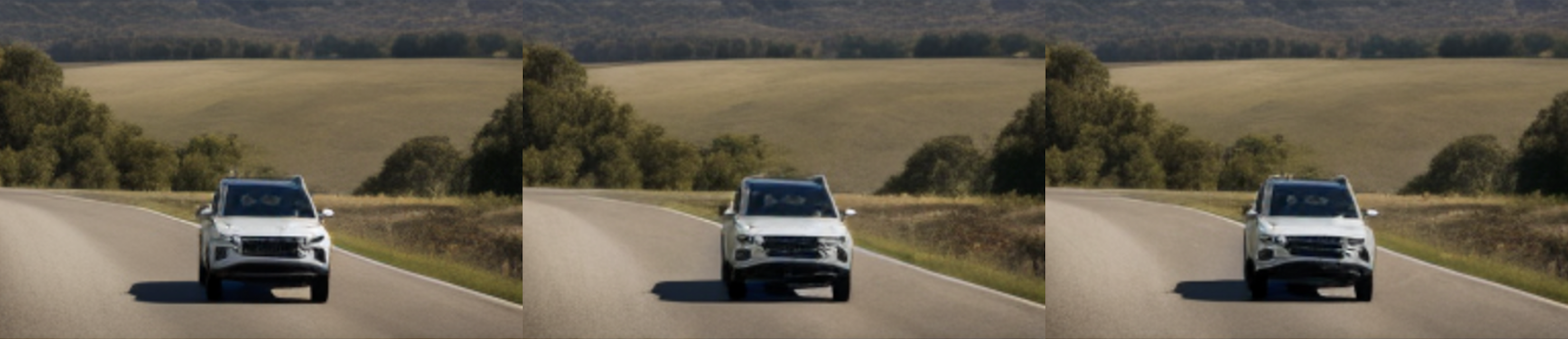}
    \caption{Frames of video generated from RFM image.}
    \label{fig:video_frames}
\end{figure}

We use an RFM image along with the positive prompt "high quality, high resolution, highly detailed, photo realistic scene" and negative prompt "low res, low resolution, poor quality, bad quality, worst quality, low quality, illustration, 3d, 2d, painting, cartoons, sketch" as input to the image-to-video model. Evaluation of the generated video is conducted by running object detection on each frame of the video as shown in \figurename \ref{fig:video_frames}. We found that object detection failures persisted in generated video frames of RFM images, implying that generated RFMs are persistent over time and realistic for video-based AV perception systems.

\subsection{Explainability Through Image-To-Text} 

Once we generate RFM images, we determine RFM cause to produce actionable insights. Since individual inspection of each RFM image is not scalable, we generate image captions using GPT4-o to identify causes of object detection failure \cite{gpt-4o_blog_post}. We provide text prompts and thermal images as inputs to the image-to-text model.

\begin{figure}
    \centering
    \includegraphics[width=\linewidth*2/3]{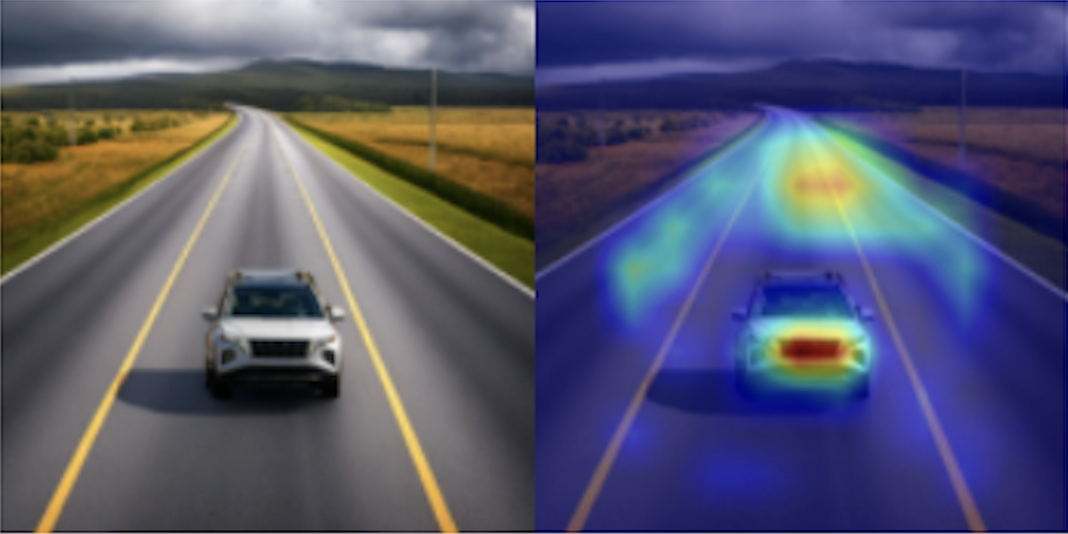}
    \caption{"The object detection model's attention appears to be overly distributed across the background scenery, especially the road and landscape, which might have misled it to misclassify or miss the object in the scene. The dispersed focus suggests the model was not sufficiently attentive to the distinct features of the vehicle, potentially causing inaccuracies in detection." captions the above RFM car image with no detections (left) based on its thermal image (right).}
    \label{fig:rfm_thermal_caption}
\end{figure}

\begin{figure}
\centering
\includegraphics[width=\linewidth]{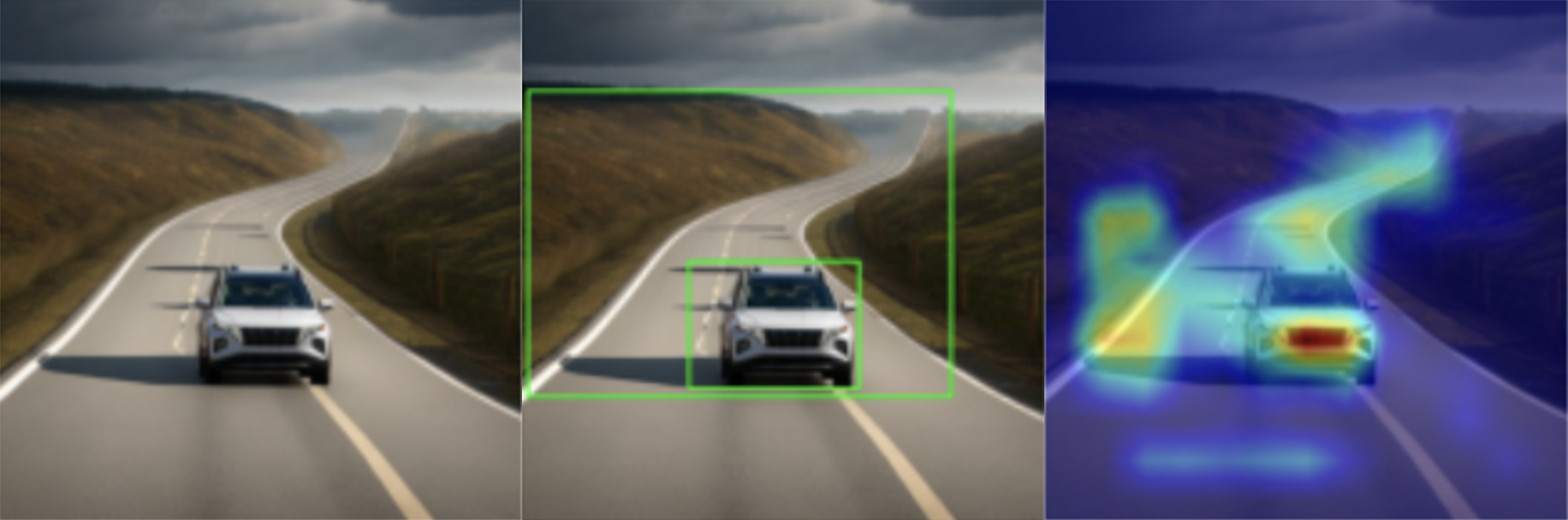}
\caption{"The model's attention appears to be drawn to portions of the backgrounds and road, which may have contributed to it missing the object. Additionally, heightened focus on areas that do not correspond to typical object features might have caused the model to hallucinate objects in the scene." captions the above RFM image with the car misclassified as a truck and a hallucinated airplane. The generated image is on the left, the overlaid detection bounding boxes in center, and the GradCAM thermal image is on the right.}
\label{fig:rfm_cause_example}
\end{figure}

We encourage the model to explain the image's content and the object detection model's performance through the system and user prompts. We use a system prompt of "You are a helpful assistant." with the below user prompt.
\begin{mdframed}[linewidth=1pt, roundcorner=5pt]
"Analyze this Grad-CAM overlay for an object detection model. 
Based on the highlighted activation zones on the vehicle and background:
- What might have misled the model to misclassify the object in the scene? If there is no evidence to answer, omit the answer from your response.
- What might have misled the model to missing the object in the scene? If there is no evidence to answer, omit the answer from your response.
- What might have misled the model to hallucinate objects in the scene? If there is no evidence to answer, omit the answer from your response.
For the questions above, provide your reasoning. 
Do not directly refer to the thermal overlay, instead refer to the model's attention. 
Your answer should be in paragraph form and one to three sentences in length. 
Your answer will be used as a caption to accompany the original image and explain the object detection model's performance. 
Viewers will not be able to see the thermal overlay, and will only see the original image and your caption. 
Your response should use clear and concise wording."
\end{mdframed}
We designed the prompts to be image-type-agnostic, allowing the same prompt to be used for caption generation for RFM and non-RFM images. We chose to withhold the results of object detection from the image-to-text model to reduce bias, prioritize explanations based on evidence from the thermal image, and eliminate possible hallucinations catered to the detection outcome. Using our system and user prompts, we generate captions for authentic, non-RFM, and RFM images, as shown in \figurename \ref{fig:rfm_thermal_caption} and \figurename \ref{fig:rfm_cause_example}.

\section{Results and Discussion}\label{sec:results-discussion}

In this section, we evaluate the performance of the Authentication framework for generating RFMs in autonomous vehicle (AV) perception systems. Our results indicate that the adversarially guided diffusion model generates rare failure images that are both realistic and capable of triggering object detection errors, such as hallucinations and misdetections.

We discovered some interesting findings: the RFMs we generated reveal vulnerabilities in the object detection model that are typically hard to catch with traditional methods (see Section \ref{subsec:RFM_Generation}). In addition, the generated captions, using our explainability tool, effectively explain what might be causing these errors, especially by identifying areas where the model may have lost focus (see Section \ref{subsec:image_to_text}). We also verified that these RFMs are both robust and consistent using our verification framework. The following subsections provide a more detailed analysis of these findings.

\subsection{RFM Generation}
\label{subsec:RFM_Generation}
The Stable Diffusion XL model was able to generate high-quality, realistic object detection failure modes for a variety of environmental conditions. Many of the generated images feature the primary object centered in the frame driving on a road with the side of the road, background, and sky visible in the image. While the majority of generated images showed natural backgrounds, uneven road markings or unnatural text or objects were included in some generated images. Induced object detection failures included missing the object of interest, hallucinating additional objects, and misclassifying the objects of interest. 6.58\% of generated images caused object detection failures to occur.

Image similarity results between the seed car image (\figurename \ref{fig:sdxl_environments}) and 76 generated adversarial images are shown in Table \ref{rfm_metrics_table}. There was no significant difference between RFM and non-RFM images in our perceptual and pixel-level similarity metrics, supporting that RFM images are not highly unrealisitic compared to non-RFM images.

\begin{table}[!t]
\renewcommand{\arraystretch}{1.3}
\caption{Object Detection and Image Similarity Metrics for Generated Car Images}
\label{rfm_metrics_table}
\centering
\begin{threeparttable}
\begin{tabular}{c c c c c c}
\hline
& \bfseries Confidence & \bfseries SSIM & \bfseries LPIPS & \bfseries MSE & \bfseries PSNR\\
\hline
Mean    & 0.9133 & 0.2337 & 0.6403 & 12.6496 & 0.05478\\
SD      & 0.0696 & 0.0151 & 0.0192 & 0.5627 & 0.0072\\
Min     & 0.5833 & 0.1918 & 0.5792 & 11.4193 & 0.04170\\
25\%    & 0.9014 & 0.2247 & 0.6305 & 12.3235 & 0.0494\\
75\%    & 0.9563 & 0.2436 & 0.6512 & 13.0617 & 0.0586\\
Max     & 0.9937 & 0.2594 & 0.6871 & 13.7989 & 0.0721\\
\hline
\end{tabular}
\begin{tablenotes}
\item{\bfseries Note:} This table shows average object-detection confidence alongside pixel-level and perceptual similarity metrics that compare each generated RFM image to the original “seed” image. Confidence is the bounding-box confidence of the correct car detection. SSIM (Structural Similarity) and LPIPS (Learned Perceptual Image Patch Similarity) measure how visually similar two images are. MSE (Mean Squared Error) and PSNR (Peak Signal-to-Noise Ratio) measure pixel-level difference between two images. Higher SSIM or PSNR means the generated image is more similar to the seed image; higher LPIPS or MSE indicates greater difference.
\end{tablenotes}
\end{threeparttable}
\end{table}

\subsection{Image-To-Text Explainable Caption Generation}
\label{subsec:image_to_text}
Captions are evaluated based on whether they identify object detection failures, whether RFM causes are correctly described, and whether the caption content accurately describes the Grad-CAM image. A caption contains a true positive if it identifies an object detection failure correctly, a true negative if it does not mention an absent object detection failure, a false positive if it falsely provides evidence for a non-existing detection failure, and a false negative if it does not mention a present detection failure. Since every caption was evaluated manually, we evaluated a maximum of 25 random car images from each category of false positive, false negative, and misclassified RFM; non-RFM; and authentic images. Due to the smaller number of generated truck images, we sampled a maximum of 10 random truck images from each category, resulting in an evaluation set of 123 car images and 23 truck images.

\begin{table*}[!t]
\renewcommand{\arraystretch}{1.3}
\caption{Quality of Generated Caption Evidence on 123 Car Images}
\label{caption_metrics_car_table}
\centering
\begin{threeparttable}
\begin{tabular}{l c c c c c c c c}
\hline
\multicolumn{1}{c}{} & \multicolumn{2}{c}{\bfseries Accuracy} & \multicolumn{2}{c}{\bfseries Recall} & \multicolumn{2}{c}{\bfseries Precision} & \multicolumn{2}{c}{\bfseries F1-Score}\\
\cline{2-9}
\multicolumn{1}{c}{} & \textit{\bfseries Mean} & \textit{\bfseries SD} & \textit{\bfseries Mean} & \textit{\bfseries SD} & \textit{\bfseries Mean} & \textit{\bfseries SD} & \textit{\bfseries Mean} & \textit{\bfseries SD}\\
\hline
All Images      & 31.44\% & \textit{0.23} & 75.26\% & \textit{0.38} & 24.17\% & \textit{0.25} & 18.30\% & \textit{0.06}\\
Generated       & 34.33\% & \textit{0.23} & 74.19\% & \textit{0.38} & 28.51\% & \textit{0.25} & 20.60\% & \textit{0.06}\\
Non-generated   & 18.84\% & \textit{0.17} & \bfseries 100.00\% & \textit{0.00} & 6.67\% & \textit{0.13} & 6.25\% & \textit{0.08}\\
RFMs            & 38.68\% & \textit{0.21} & 75.26\% & \textit{0.38} & 36.87\% & \textit{0.22} & 24.75\% & \textit{0.61}\\
FN RFMs         & 33.33\% & \textit{0.20} & 65.00\% & \textit{0.42} & 35.45\% & \textit{0.23} & 22.94\% & \textit{0.05}\\
FP RFMs         & 41.98\% & \textit{0.19} & 66.67\% & \textit{0.34} & \bfseries 45.90\% & \textit{0.24} & 27.18\% & \textit{0.07}\\
Miscl RFMs      & \bfseries 47.22\% & \textit{0.21} & 92.31\% & \textit{0.22} & 40.00\% & \textit{0.17} & \bfseries 27.91\% & \textit{0.06}\\
non-RFMs        & 17.46\% & \textit{0.18} &-&-&-&-&-&-\\
\hline
\end{tabular}
\begin{tablenotes}
\item{\bfseries Note:} This table shows caption performance on all images, generated and non-generated images, all RFM images, images with missed detections (FN RFMs), images with hallucinations (FP RFMs), images with misclassifications (Miscl RFMs), and non-RFM images with a car as the object of interest.
\end{tablenotes}
\end{threeparttable}
\end{table*}

\begin{table*}[!t]
\renewcommand{\arraystretch}{1.3}
\caption{Quality of Generated Caption Evidence on 24 Truck Images}
\label{caption_metrics_truck_table}
\centering
\begin{threeparttable}
\begin{tabular}{l c c c c c c c c}
\hline
\multicolumn{1}{c}{} & \multicolumn{2}{c}{\bfseries Accuracy} & \multicolumn{2}{c}{\bfseries Recall} & \multicolumn{2}{c}{\bfseries Precision} & \multicolumn{2}{c}{\bfseries F1-Score}\\
\cline{2-9}
\multicolumn{1}{c}{} & \textit{\bfseries Mean} & \textit{\bfseries SD} & \textit{\bfseries Mean} & \textit{\bfseries SD} & \textit{\bfseries Mean} & \textit{\bfseries SD} & \textit{\bfseries Mean} & \textit{\bfseries SD}\\
\hline
All Images      & 25.00\% & \textit{0.20} & 72.73\% & \textit{0.40} & 13.56\% & \textit{0.18} & 11.43\% & \textit{0.06}\\
Generated       & \bfseries 28.07\% & \textit{0.20} & 70.00\% & \textit{0.44} & 15.56\% & \textit{0.19} & 12.73\% & \textit{0.06}\\
Non-generated   & 13.33\% & \textit{0.16} & \bfseries 100.00\% & \textit{0.00} & 7.14\% & \textit{0.13} & 6.67\% & \textit{0.13}\\
RFMs            & 27.27\% & \textit{0.19} & 72.73\% & \textit{0.40} & \bfseries 27.59\% & \textit{0.18} & \bfseries 20.00\% & \textit{0.03}\\
non-RFMs        & 23.08\% & \textit{0.20} &-&-&-&-&-&-\\
\hline
\end{tabular}
\begin{tablenotes}
\item{\bfseries Note:} This table shows caption performance on all images, generated and non-generated images, RFM images, and non-RFM images with a truck as the object of interest. 
\end{tablenotes}
\end{threeparttable}
\end{table*}

We seek to determine if the image-to-text model can explain the content of authentic images used as seed images for the inpainting model, creating a baseline for comparison to generated image captions. We use the 6 car images and 5 truck images used as seed images for the inpainting model supplemented by 17 car video frames as our baseline. Tables \ref{caption_metrics_car_table} and \ref{caption_metrics_truck_table} show the generated caption results for car and truck images, respectively. The captions suggested that misclassification was likely in all but one non-generated car image when none were misclassified. The generated captions identified and provided plausible evidence for present hallucinations. The image-to-text model's performance on authentic, non-generated images shows a propensity to assume object detection failure, especially misclassification.

We generate captions for non-RFM images to explain why these images did not cause an object detection failure and find cases where the environment disrupted model performance without a full detection failure. We use 42 non-RFM car images and 13 non-RFM truck images, including generated and authentic images, as our test set. The generated captions for non-RFM images overemphasize the potential for object detection failures, resulting in low accuracy as shown in Tables \ref{caption_metrics_car_table} and \ref{caption_metrics_truck_table}. The true negative rate (TNR) was 17.46\% for car image captions and 23.08\% for truck image captions. The image-to-text model's performance on non-RFM images is hampered by frequent false positives.

We generate captions for RFM images to increase interpretability and accessibility by explaining potential reasons for object detection failure. We generate captions for 81 RFM car images which induce 46 false negatives, 27 false positives, and 24 misclassifications as our car test set. We generate captions for 11 RFM truck images. Due to this small set of RFM truck images, we will not break the truck test set into subsets for analysis. The generated captions for RFM images tended to include superfluous evidence rather than exclude relevant evidence, leading to lower recall relative to precision. We observe greatest overall caption performance on misclassified images and least superfluous evidence for images with hallucinations. Evidence for missed detections was most likely to be overlooked, leading to degraded caption performance. The quantitative results for car and truck RFM captions are shown in Tables \ref{caption_metrics_car_table} and \ref{caption_metrics_truck_table}.

We determine whether the image-to-text model is able to accurately describe the images and provide coherent evidence for detection errors. The generated captions accurately describe the image and thermal map in 99.19\% of 123 car images and 100\% of 24 truck images and provide accurate evidence for identified detection errors for 97.44\% of 81 car RFM images and 100\% of 11 truck RFM images. The generated captions consistently describe both car and truck images, indicating that the image-to-text model is able to identify the visual features needed to identify RFM causes. Further, the generated captions are able to describe potential causes for identified RFM errors, detailing areas where the object detection model's attention was disrupted.

\subsection{RFM Causes}
\label{subsec:RFM_Causes}

We observe that disrupted attention is a common shared quality between RFM images. Images that do not trigger a detection error tend to be high-contrast and include bright green foliage, while RFM images often have low contrast, occluded vision, or autumn foliage. We evaluate RFM car images generated from the seed image in \figurename \ref{fig:sdxl_environments}, sampling a maximum of 25 images from each category of false positive, false negative, and misclassified RFMs. In RFM images with missed detections, the attended areas of the image tend to stray from the primary object as shown in \figurename \ref{fig:rfm_thermal_caption}. In 46 car images with false negative detections, the detector's focus is most often on the road while rarely straying to the sky. We observed low-contrast images with fog or rain tend to cause disrupted attention that leads to missed detections. For RFM images with hallucinations, there tends to be a misplaced focus on the area where the hallucination is detected. In 27 car images with hallucinations, the area with most focus was fairly evenly distributed across images, indicating that any area of the image can trigger a hallucination. We observed trees and signs on the side of the road or in the background as common causes of hallucinations. For misclassified RFM images, attention is often focused on one aspect of the primary object or is overly dispersed. In 24 car images with misclassifications, the detector's focus was most often deflected to the road. A common misclassification case occurs when part of the road can be seen behind the roof of a car, leading the detector to misclassify it as a truck as shown in \figurename \ref{fig:rfm_cause_example}. Table \ref{reason_metrics_car_table} shows the percentage of RFM images where attention was diverted to specific image regions.

\begin{table}[!t]
\renewcommand{\arraystretch}{1.3}
\caption{Areas of Disrupted Attention in RFM Images}
\label{reason_metrics_car_table}
\centering
\begin{threeparttable}
\begin{tabular}{c c c c c}
\hline
\bfseries Important Area & \bfseries FN & \bfseries FP & \bfseries Miscl & \bfseries All\\
\hline
Road            & \bfseries 67.39\% & 39.29\% & \bfseries 75.00\% & \bfseries 60.56\%\\
Side of Road    & 23.91\% & \bfseries 42.86\% & 20.83\% & 29.20\%\\
Background      & 32.61\% & 39.29\% & 37.50\% & 36.46\%\\
Sky             & 6.52\% & 28.57\% & 16.67\% & 17.25\%\\
Unnatural Object& 0.00\% & 3.57\% & 8.33\% & 3.97\%\\
\hline
\end{tabular}
\begin{tablenotes}
\item{\bfseries Note:} This table shows areas of the image to which detector attention was disrupted when missed detections (FN), hallucinations (FP), and misclassifications (Miscl), as well as for all RFM images.
\end{tablenotes}
\end{threeparttable}
\end{table}

\section{Future Work}\label{sec:future-work}

Light Detection and Ranging (LiDAR) data can be integrated into the existing pipeline to leverage 3D information and improve system robustness. The idea is to process LiDAR point clouds to generate segmentation maps that complement image-based analysis by including downsampling, and projecting segments onto the image plane. These masks can then be applied to adversarial guidance and failure mode analysis, with modifications ensuring that LiDAR-generated masks improve object detection and consistency checks. In future work, we could extend the pipeline to handle both image and LiDAR data simultaneously.

We demonstrate that RFM images altered through null text inversion maintain their object detection failures. In future work, this technique could be applied as a hypothesis-testing tool to verify and clarify RFM causes and bounds.

Current generated captions accurately describe the image and attention heatmap but overestimate the number of object detection failures. The current image-to-text implementation is image-type-agnostic, allowing caption generation with the same prompt across authentic, non-RFM, and RFM images and leaving the model unbiased from the object detection results. Alternatively, the captions could focus on explaining the confirmed object detection failures by providing the image-to-text model information about the number of detections, detection confidence, or bounding boxes. Another approach is to present the image-to-text model with sample answers, such as ideal captions for it to mimic or example Grad-CAM images matched with the present object detection failures. Further, asking the model to include answers with evidence rather than exclude answers without evidence might lead to more precise captions with fewer false positives.

Textual inversion allows one pseudo-word to represent a complex, personalized concept by training the embedding space of a text-to-image diffusion model. This token can then be used to prompt the diffusion model for custom, personal images \cite{textualinversion_paper}. Using captions generated by the image-to-text model, we could collect a set of RFM images with one complex RFM cause in common, for example RFMs caused by yellow foliage in the background and water on the road. Using textual inversion, a specific pseudo-word can represent the RFM cause in the image set. By prompting the Stable Diffusion model with the customized token, we could generate more specific RFM images without relying on long textual descriptions. We could also use this method to test hypotheses about RFM cause, validating the image-to-text model's generated captions.

Shaham et al. uses a pretrained vision-language model equipped with tools to iteratively experiment on a target model's subcomponents to explain its behavior \cite{maia_paper}. In future work, a similar framework could be implemented to generate and test RFM cause hypotheses using our custom toolbox of text-to-image, null text inversion, text-to-image, and image-to-video methods.

\section{Conclusion}\label{sec:conclusion}

In this paper, we tackled a major gap in autonomous vehicle testing by generating realistic adversarial images leveraging stable diffusion in-painting models. The denoising process in stable diffusion was driven by the object detection model’s loss function, preserving the object of interest (i.e., a car) but generating a new environment that tends to fool the object detection model. In addition, we provide an explainable method that offers a fuller picture of why and how these RFMs cause the object detection model to be fooled. As we move forward, combining adversarial methods with clear explanations and diverse data sources will be vital for building safer, more trustworthy self-driving systems.

\section*{Acknowledgment}
This work was supported by the MITRE Independent Research and Development Program. The authors would also like to thank Dr. Scott Rosen and Anna Raney for their valuable peer reviews and feedback, which greatly enhanced the quality of this work.

\newpage


\begin{thebibliography}{10}

\bibitem{WANG202417}
H.~Wang, W.~Shao, C.~Sun, K.~Yang, D.~Cao, and J.~Li, ``A survey on an emerging
  safety challenge for autonomous vehicles: Safety of the intended
  functionality,'' {\em Engineering}, vol.~33, pp.~17--34, 2024.

\bibitem{Wang_LongTail2022}
J.~Wang, X.~Wang, T.~Shen, Y.~Wang, L.~Li, Y.~Tian, H.~Yu, L.~Chen, J.~Xin,
  X.~Wu, N.~Zheng, and F.-Y. Wang, ``Parallel vision for long-tail
  regularization: Initial results from ivfc autonomous driving testing,'' {\em
  IEEE Transactions on Intelligent Vehicles}, vol.~7, no.~2, pp.~286--299,
  2022.

\bibitem{sd2.0_paper}
R.~Rombach, A.~Blattmann, D.~Lorenz, P.~Esser, and B.~Ommer, ``High-resolution
  image synthesis with latent diffusion models,'' in {\em Proceedings of the
  IEEE/CVF Conference on Computer Vision and Pattern Recognition (CVPR)},
  pp.~10684--10695, June 2022.

\bibitem{fasterrcnn}
S.~Ren, K.~He, R.~Girshick, and J.~Sun, ``Faster r-cnn: Towards real-time
  object detection with region proposal networks,'' {\em IEEE Transactions on
  Pattern Analysis and Machine Intelligence}, vol.~39, no.~6, pp.~1137--1149,
  2017.

\bibitem{usdotav_website}
{US Department of Transportation}, ``Automated vehicles comprehensive plan.''
\newblock (accessed December 4, 2024).

\bibitem{Liu2024}
H.~X. Liu and S.~Feng, ``Curse of rarity for autonomous vehicles,'' {\em Nature
  Communications}, vol.~15, p.~4808, Jun 2024.

\bibitem{ho2020denoisingdiffusionprobabilisticmodels}
J.~Ho, A.~Jain, and P.~Abbeel, ``Denoising diffusion probabilistic models,''
  2020.

\bibitem{advdiffuser_paper}
X.~Chen, X.~Gao, J.~Zhao, K.~Ye, and C.-Z. Xu, ``{AdvDiffuser: Natural
  Adversarial Example Synthesis with Diffusion Models},'' in {\em {2023
  IEEE/CVF International Conference on Computer Vision (ICCV)}}, pp.~01--06,
  IEEE.

\bibitem{kirillov2023segment}
A.~Kirillov, E.~Mintun, N.~Ravi, H.~Mao, C.~Rolland, L.~Gustafson, T.~Xiao,
  S.~Whitehead, A.~C. Berg, W.-Y. Lo, P.~Dollár, and R.~Girshick, ``Segment
  anything,'' 2023.

\bibitem{segment_anything_paper}
A.~Kirillov, E.~Mintun, N.~Ravi, H.~Mao, C.~Rolland, L.~Gustafson, T.~Xiao,
  S.~Whitehead, A.~C. Berg, W.-Y. Lo, P.~Doll{\ifmmode\acute{a}\else\'{a}\fi}r,
  and R.~Girshick, ``{Segment Anything},'' {\em arXiv}, Apr. 2023.

\bibitem{sdxl_paper}
D.~Podell, Z.~English, K.~Lacey, A.~Blattmann, T.~Dockhorn,
  J.~M{\ifmmode\ddot{u}\else\"{u}\fi}ller, J.~Penna, and R.~Rombach, ``{SDXL:
  Improving Latent Diffusion Models for High-Resolution Image Synthesis},''
  {\em arXiv}, July 2023.

\bibitem{latent_diffusion_models_paper}
R.~Rombach, A.~Blattmann, D.~Lorenz, P.~Esser, and B.~Ommer, ``{High-Resolution
  Image Synthesis with Latent Diffusion Models},'' {\em arXiv}, Dec. 2021.

\bibitem{sd-nae_paper}
Y.~Lin, J.~Zhang, Y.~Chen, and H.~Li, ``{SD-NAE: Generating Natural Adversarial
  Examples with Stable Diffusion},'' {\em arXiv}, Nov. 2023.

\bibitem{diffusion_models_counterfactual_explanations_paper}
G.~Jeanneret, L.~Simon, and F.~Jurie, ``{Diffusion Models for Counterfactual
  Explanations},'' {\em arXiv}, Mar. 2022.

\bibitem{rombach2022highresolutionimagesynthesislatent}
R.~Rombach, A.~Blattmann, D.~Lorenz, P.~Esser, and B.~Ommer, ``High-resolution
  image synthesis with latent diffusion models,'' 2022.

\bibitem{gpt-4o_blog_post}
OpenAI, ``{Hello GPT-4o}.''
\newblock (accessed December 5, 2024).

\bibitem{grad-cam_paper}
R.~R. Selvaraju, M.~Cogswell, A.~Das, R.~Vedantam, D.~Parikh, and D.~Batra,
  ``Grad-cam: Visual explanations from deep networks via gradient-based
  localization,'' {\em International Journal of Computer Vision}, vol.~128,
  p.~336–359, Oct. 2019.

\bibitem{null-text_inversion_paper}
R.~Mokady, A.~Hertz, K.~Aberman, Y.~Pritch, and D.~Cohen-Or, ``{Null-text
  Inversion for Editing Real Images using Guided Diffusion Models},'' {\em
  arXiv}, Nov. 2022.

\bibitem{consistI2V_paper}
W.~Ren, H.~Yang, G.~Zhang, C.~Wei, X.~Du, W.~Huang, and W.~Chen, ``{ConsistI2V:
  Enhancing Visual Consistency for Image-to-Video Generation},'' {\em arXiv},
  Feb. 2024.

\bibitem{textualinversion_paper}
R.~Gal, Y.~Alaluf, Y.~Atzmon, O.~Patashnik, A.~H. Bermano, G.~Chechik, and
  D.~Cohen-Or, ``An image is worth one word: Personalizing text-to-image
  generation using textual inversion,'' {\em arXiv}, 2022.

\bibitem{maia_paper}
T.~R. Shaham, S.~Schwettmann, F.~Wang, A.~Rajaram, E.~Hernandez, J.~Andreas,
  and A.~Torralba, ``A multimodal automated interpretability agent,'' 2024.

\end{thebibliography}

\end{document}